\pdfoutput=1

\documentclass{article}
\usepackage{ijcai18}

\usepackage{times}
\usepackage{xcolor}
\usepackage{soul}
\usepackage[utf8]{inputenc}
\usepackage[small]{caption}
\usepackage{amsmath}
\usepackage{amssymb}
\usepackage[draft]{todonotes}  
\usepackage[font=small,skip=0pt]{caption}

\usepackage{graphicx}

\makeatletter
\newcommand*{\centerfloat}{%
  \parindent \z@
  \leftskip \z@ \@plus 1fil \@minus \textwidth
  \rightskip\leftskip
  \parfillskip \z@skip}
\makeatother

\title{Causal Learning and Explanation of Deep Neural Networks via Autoencoded Activations}

\author{
Michael Harradon$^1$,
Jeff Druce$^1$,
Brian Ruttenberg$^1$
\\
$^1$ Charles River Analytics, Cambridge, MA\\
\{mharradon, jdruce, bruttenberg\}@cra.com
}

\begin{document}

\maketitle

\begin{abstract}
Deep neural networks are complex and opaque. As they enter application in a variety of important and safety critical domains, users seek methods to explain their output predictions. We develop an approach to explaining deep neural networks by constructing causal models on salient concepts contained in a CNN. We develop methods to extract salient concepts throughout a target network by using autoencoders trained to extract human-understandable representations of network activations. We then build a bayesian causal model using these extracted concepts as variables in order to explain image classification. Finally, we use this causal model to identify and visualize features with significant causal influence on final classification.
\end{abstract}

\section{Introduction}
\label{intro}
Artificial intelligence (AI)-enabled systems are playing an increasingly prominent role in our society and significantly disrupting how commercial and government organizations operate. In particular, the rate of advances in machine learning (ML) using Deep Neural Networks (DNNs) is staggering, enabling AI-enabled systems to master complex tasks including Go \cite{alphago}, autonomous driving \cite{nvidiadriving}, and even predicting census data \cite{gebru2017fine}. These incredible advances, however, have come at a significant cost: DNNs are complex, opaque, and derive their power from millions of parameters that must be trained on large data sets. Thus, understanding or explaining \textit{why} a DNN came to a particular conclusion is a difficult task. Yet as AI-enabled systems become more prevalent in our lives, this lack of explainability and intepretability comes with serious societal consequences. For example, ML is now being used to predict recidivism in the criminal justice system \cite{berk2017impact}, interventions in the child welfare system \cite{cuccaro2017risk}, and cancer from radiology images \cite{esteva2017dermatologist}. Lack of explainability in these applications can have life or death consequences. 

The need for explainability and interpretability has certainly not gone unnoticed in the AI and social science communities. Important strides are being made to either imbue DNNs with inherent explainability or to achieve the predictive power of DNNs using more explainable methods. In particular, there has been a significant amount of progress explaining Convolutional Neural Networks (CNNs) in the image domain, most likely due to the ease of visualizing explanations. For example, GradCAM \cite{selvaraju2016grad} and LRP \cite{binder2016layer} are two popular methods of generating saliency maps that indicate the relevance of pixels in the image to the output of the CNN. 

While these methods have vastly improved the explainability landscape, they lack one critical element needed for true understanding by humans: They have limited causal interpretations. Causality as a means of explanation has deep roots within the AI community \cite{pearl2009causality}. A \textit{causal} explanation of a DNN's operation provides the end--consumer of the DNN output (i.e., a human) with the understanding of \textit{how} and \textit{what} can be changed about a DNN or its input that results in an impactful change in the output.  In highly sensitive domains like credit scoring or money laundering, a causal explanation is critical as system implementers must justify the operation of their ML models to government regulators. One major challenge, however, is that any causal explanation must be formulated in terms of concepts or variables that are understandable to a human; otherwise the explanation may end up as obfuscated as the original model.

The hypothesis that we explore in this paper posits that a human--understandable causal model of the operation of a DNN that allows for arbitrary causal interventions and queries is an effective, and often necessary, tool for explainability and interpretability. Arbitrary interventions allows the user to understand the chain of causal effects from DNN input, to low--level features of the domain, to high--level human--understandable concepts, to DNN outputs. Critically, such a model, if constructed accurately, could support introspective queries on a DNN not supported by other methods such as \textit{counterfactual} queries. For example, one might be able to ask "What is the probability the car would have turned right had there not been a pedestrian present in the input image?" Such a capability is a powerful tool for debugging, understanding bias, and ensuring safe operation of AI systems. 

In this paper, we explore this hypothesis and demonstrate that a causal approach to DNN explanation is possible and yields valuable information about various classification systems. As part of this approach, we extract low--dimensional concepts from DNNs to generate a human--understandable ``vocabulary'' of variables. We then perform interventional experiments to learn a graphical causal model (i.e., Bayesian network \cite{pearl2009causality}) that relates the DNN's inputs to the concepts, and the concepts to the DNN's outputs. Finally, we demonstrate the explanatory power of such a model by identifying the concepts in several networks with the highest expected causal effect on the outputs. Our contributions can be summarized as follows:

\begin{itemize}
\item A causal model of a DNN's operation formulated in human--understandable concepts to aid explainability
\item An unsupervised technique to extract concepts from DNNs that have high likelihood of being human--understandable
\item A proposed way to measure the causal effects of inputs and concepts on a DNN's outputs
\end{itemize}

The remainder of this paper is organized as follows. In Section~\ref{related} we discuss related explainability work. In Section~\ref{causality} we formulate the notions of causality in DNNs. We discuss the extraction of human--understandable concepts from DNNs in Section~\ref{extraction}. We run several experiments and present examples of our results in Section~\ref{results}, and finally conclude in Section~\ref{conclusion}

\section{Related Work}
\label{related}
Many recent works employ saliency for visual explanations. Several previous works have visualized predictions by emphasizing pixels which have influential values (i.e., if the pixel values change significantly, the output of the network also changes significantly). In an early work in this domain \cite{erhan2009visualizing}, an input image which maximally activated a neuron of interest was found by gradient ascent in the image domain; this work was extended in \cite{simonyan2013deep} to obtain class-specific saliency maps. Beyond manipulating the image space and monitoring the impact on the output, other works have considered an analysis on the learned \emph{features} of the network in order to glean understanding on how the network functions. In the seminal work by \cite{zeiler2014visualizing}, a multi-layer deconvolutional network was used to project features activations (the output from convolution maps on images) back to input pixel space. Gradient information flowing into the penultimate convolutional layer was used in GradCAM \cite{selvaraju2016grad} to identify discriminating patterns in input images. 

Rather than manipulating input images and their activations within a network, methods have been explored which generate images which produce a desired response in the network; the constructed result can help explain what input has a maximal desired response in the network. Techniques have been developed to invert the network by constructing images that activate the network identically \cite{mahendran2015understanding}, or maximally \cite{erhan2009visualizing}. Fascinating images can be constructed which maximally activate deep convolutional filters in a network \cite{mordvintsev2015inceptionism}, which illustrate the ellaborate features generated by the larger deep net architectures. 

Methods which do not rely on gradients have also recently gained traction for visual explanations. Layer-Wise Relevance Propagation \cite{bach2015pixel} relies on a conservation principle to redistribute the prediction of the network backwards until a relevance score is computed for each element in input space, and has been shown to produce interpretable heatmaps to explain individual classifications \cite{binder2016layer}. The method reported in \cite{xie2017relating}, aims to relate human understandable concepts to network outputs, by employing a deconvolution and masking- based technique to find and score the strength of distributed representations of input concepts across late stage feature maps. Other methods do not directly consider the network at all, but rather locally approximate the "blackbox" models for simpler, explainable ones have shown to generate results which inspire trust in users \cite{ribeiro2016should}. 

\section{Causal Modeling}
\label{causality}

Causality has a long history in AI and numerous ML efforts have focused on building realistic and accurate causal models of the world instead of statistical models~\cite{pearl2018theoretical}. This has resulted in a plethora of causal formalisms and semantics in various fields~\cite{granger1980testing}. However, in this work, we frame the semantics of causality in terms of the interventional effects of Pearl's \textit{do}-calculus~\cite{pearl2009causality}. For a directed graphical model $G$ defining a causal diagram on a set of variables $X = {x_1, ..., x_n}$, we define the joint probability distribution over $X$ as (from Pearl Eqn 3.5)
\begin{equation}
P(x_1, ..., x_n) = \underset{i}{\prod}P(x_i|pa_i)
\end{equation}
where $pa_i$ are the parents of $x_i$. An intervention on $G$ by setting $x_i = x_i'$, denoted as $do(x_i')$, induces a modified graph $G'$ where the edges from $pa_i$ to $x_i$ are removed, resulting in the postintervention distribution (from Pearl Eqn. 3.10):
\begin{equation}
P(x_1, ..., x_n | do(x_i')) = 
\begin{cases}
\underset{j \neq i}{\prod}P(x_j|pa_j) & \mbox{if } x_i = x_i' \\
0 & \mbox{if } x_i \neq x_i' \\
\end{cases}
\end{equation}
The semantics of the causality are important for explaining DNNs because, in essence, explanations \textit{must be} causal models. That is, when one seeks an explanation of a network's decisions, one is equivalently asking "What changes can be made to the input for the output to change or stay the same?". This formulation of causation as explanation is well supported in the literature~\cite{woodward2005making}. We consider a causal model that defines the joint distribution $P(\mathbb{O}, \mathbb{P}, \mathbb{X})$ over a set of DNN outputs $\mathbb{O}$, inputs $\mathbb{P}$, and intermediate variables $\mathbb{X}$.

More importantly, the notion of interventions provides a clear and mathematically sound mechanism for the user to understand \textit{why} a DNN produces different output values. To put it another way, the explanation of an observed DNN output can be formulated as an intervention. For example, if we say ``the DNN recognized a pedestrian in the image because it saw a head'', that implies that the DNN thinks there is both a pedestrian and a head in the image. If the DNN had not detected a head, or if we intervene on the input to remove the head from the image, then the probability of detecting a pedestrian would have changed. 

Existing explanation methods for DNNs, in particular on image classification tasks, lack the ability to provide this concrete causal interpretation to the user. For example, gradient--based methods of explanation such as layerwise relevance propagation (LRP)~\cite{bach2015pixel} and Grad--CAM~\cite{selvaraju2016grad} attempt to explain the output activation of a class $C$ in terms of the input $P$ activations, realized for a specific input $P_j$. While these methods are certainly useful, they don't provide the causal intervention semantics that are sufficient for robust explanation. Due to discontinuous and saturated gradients, they only indicate causality over a restricted domain where the function defined by the network can be approximated with a linear function. Adversarial instances generated using gradient descent \cite{adversarial} provide indication that the local behavior of functions defined by trained DNNs does not have semantic relevance, which suggests that in addition to the interventions defined by gradient based methods being \textit{restricted} to a small domain, they are also \textit{uninterpretable} in that they are considering a semantically dubious aspect of DNNs. Methods like LRP avoid some of the practical issues with the gradient based methods by ``redistributing'' activation levels to the most relevant pixels, but again do not provide the explicit causal intervention semantics desired for effective explanation.
\subsection{Causal Representation in DNNs}
Given that we want a causal model that reflects the intervention semantics, the question arises as to what is represented the joint distribution $P(\mathbb{O}, \mathbb{P}, \mathbb{X})$. With full access to the internals a DNN, we already have a causal representation of the DNN in terms of its structure and weights. That is, for a given DNN, we can define $\mathbb{X} = \mathbb{R}$, where $\mathbb{R}$ is the set of all neurons in the network, and learn this joint distribution by experimentally intervening on network representations \cite{pearl2009causality}. A user could then ask counterfactual questions about the network, i.e. $P(\mathbb{O}, \mathbb{P}, \mathbb{X} | do(x_i'))$ for any input, output, or internal neuron in the network.

While this method is, on a technical level, correct, it serves poorly as a model for explanation. This is due to the lack of human--level concepts that underlie any arbitrary neuron in the network: saying neuron $r_i$ caused the network to detect a pedestrian may be technically correct but does not satisfy the needs of the eventual human that will ingest the explanations. In addition, the language of interventions that a human would use to understand the network are not represented as individual neurons. For example, the user cannot inquire about the causal impact of a head towards the detection of a pedestrian if the only method of intervention available is at neuron granularity. As a result, we posit that a DNN causal model must be constructed at a conceptual granularity that is meaningful to humans. 

We propose that a causal model for DNNs should be represented by joint distribution over $\mathbb{O}$, $\mathbb{P}$, and a set of \textit{concepts} $\mathbb{C}$. The process for deriving $\mathbb{C}$ is described by a function $\mathit{f_{\mathbb{R}}}:\mathbb{R} \rightarrow \mathbb{C}$ over a specific DNN that transforms the representation of neurons and their activations into a set of concept variables. Ideally, $\mathit{f_{\mathbb{R}}}$ would have the following properties:
\begin{align}
\int_R P(\mathbb{O}, \mathbb{P}, \mathbb{R}) & = \int_C P(\mathbb{O}, \mathbb{P}, \mathbb{C}) \\
P(\mathbb{O}, \mathbb{P}| R, do(p_i')) & = P(\mathbb{O}, \mathbb{P}| C, do(p_i'))
\end{align}
That is, we want the joint distribution over the inputs and outputs to be the same for both the neuron--level and concept--level causal models. Furthermore, we also want the same causal dependencies to hold with respect to input interventions. The semantics of $\mathbb{C}$ are open and can range from simple groups of neuron to high--level human concepts like arms, legs, and heads. This subjectivity in the concepts is quite powerful, as it presents a method to explain DNN operation without compromising the true causal semantics of the network, and provides the ability to allow users propose human--understandable interventions and queries. 
\subsection{Computing Causal Effects}
Given a causal model defined by $P(\mathbb{O}, \mathbb{P}, \mathbb{C})$, there are a number of interesting queries one might ask to better understand and explain the operation of a DNN. 
In this work, we propose a measure we call the \textit{expected causal effect}. First, we define the causal effect of an intervention on $x_i'$ on $X_j =  x_j$ given any evidence $Z$ as:
\begin{equation}
\mathrm{Effect}(x_i \rightarrow x_j, Z) =  P(x_j|do(x_i'), Z_{X_i}) - P(x_j| Z_{X_i})
\end{equation}
where $Z_{X_i}$ is all evidence that is not a descendant of $X_i$. This definition is similar to traditional measures of causal effect~\cite{rosenbaum1983central} except the key difference here is that we are comparing the effect of intervention on $X_i$ to no intervention at all. Given the effect, we then define the expected causal effect as:
\begin{align}
\begin{split}
E_{X_i} & [\mathrm{Effect}(x_i \rightarrow x_j, Z)]  = \\
& \sum_{x_i \in X_i} P(X_i = x_i | Z) \,\, \mathrm{Effect}(x_i \rightarrow x_j, Z)
\end{split}
\end{align}
Note that to compute the expectation, we use all of the evidence $Z$, as we only want to consider effects on outcomes that are possible given the evidence. For example, if we observe the output of a DNN is true, then the causal effect of any variable on a false output is always zero (for a binary DNN output). Using this formulation, we have a simple yet effective measure to quantify the impact of various DNN inputs and concepts on outputs.

\section{Concept Extraction}
\label{extraction}
In order to constuct our causal model we would first like to create a set of interpretable concepts $C$ that satisfy the above causal semantics. One way to construct concepts that satisfy these semantics would be to consider network activations. The goal would then be to learn a causal model relating these activations. The concepts that one chooses, however, do not need to be restricted to those represented explicitly by the network activations to properly satisfy the semantics. As a simple example, one could consider inserting two linear dense layers in a deep neural network such that their weight matrices $W_1$ and $W_2$ were inverses of each other. The activations after multiplication by $W_1$ could take the form of any linear combination of the prior activations, yet the final network output would be unaffected. So the specific representation of instance features given by activation values does not necessarily have any special relevance. We instead choose to find a concept representation, that is, a transformation on the activations, that's maximally interpretable. In addition to satisfying the causal intervention criteria described in section 3, these interpretable concepts should satisfy a few additional criteria:

\begin{enumerate}
\item Concepts should be low-dimensional to minimize the amount of investigation a human would need to employ.
\item Concepts should be interpretable - in the case of images, we would like activations to be restricted to contiguous areas containing consistent, interpretable visual features.
\item Concepts should contain all of the relevent information needed for achieving the target network's task (image classification in the cases we consider).
\end{enumerate}

We create an auxilliary neural network model that constructs concept representations satisfying these properties through training with a specially designed loss function. Specifically, the form of this model would be that of an autoencoder, whereby specific compression and interpretability losses could be applied, and retention of information required for classification can be ensured by the application of reconstruction losses. This approach has been recently employed to construct interpretable representations of learned features used in classification \cite{qi2017embedding}. Our approach differs in two main ways. First, rather than training the autoencoder to match output classification based on a linear function of our coded features, we employ two different reconstruction losses in our autoencoder. Second, we train multiple autoencoders throughout the deep neural network to construct interpretable representations of the activations throughout the network.

To elaborate on the loss function we employ, a "shallow" reconstruction loss is applied that is simply the $L_1$ norm of the difference between the input and the output activations of our autoencoder. We also employ a "deep" reconstruction loss that ensures that reconstructed activations result in the same classification output after being passed through the rest of the network. This loss takes the form of the KL-divergence between the output probability distributions for the original network and copy of the network with an autoencoder inserted at a given activation. That is, for the target activations $a_i$, coding function $c_\theta$, decoding function $d_\theta$, and function $r$ describing the application of the rest of the network layers following the autoencoded activation layer:
\begin{align}
L_{shallow}(\theta;a_i) &= | d_\theta (c_\theta(a_i)) - a_i |_1 \\
L_{deep}(\theta;a_i) &= KL(r(a_i)||r(d_\theta (c_\theta(a_i))))
\end{align}

The deep reconstruction loss is enforced much more strongly with a large loss weighting hyperparameter $\lambda_{deep} >> \lambda_{shallow}$. This allows our decoder to reconstruct slightly altered network activations while ensuring that important downstream network activations are unaffected. The added flexibility of fitting the deep loss instead of the shallow loss enables our autoencoder to learn more interpretable representations by relaxing the requirement that activations are decoded precisely. We additionally apply an "interpretability" loss which serves to mathematically quantify properties that one would associate with interpretable concept images, much like prior work \cite{qi2017embedding} - in particular we employ a sparsity loss, a cross-entropy loss and a total-varation loss to encourage spatially smooth, independent coded concept features. Our total autoencoder loss is then:
\begin{align}
\begin{split}
L(\theta;x_i) = & \lambda_{shallow} L_{shallow}(\theta;x_i) + \\
& \lambda_{deep} L_{deep}(\theta;x_i) + \\
& \lambda_{interpretability} L_{interpretability} (\theta;x_i)
\end{split}
\end{align}

\begin{figure}[h!]
  \vspace{-10pt}
  \includegraphics[width=\linewidth]{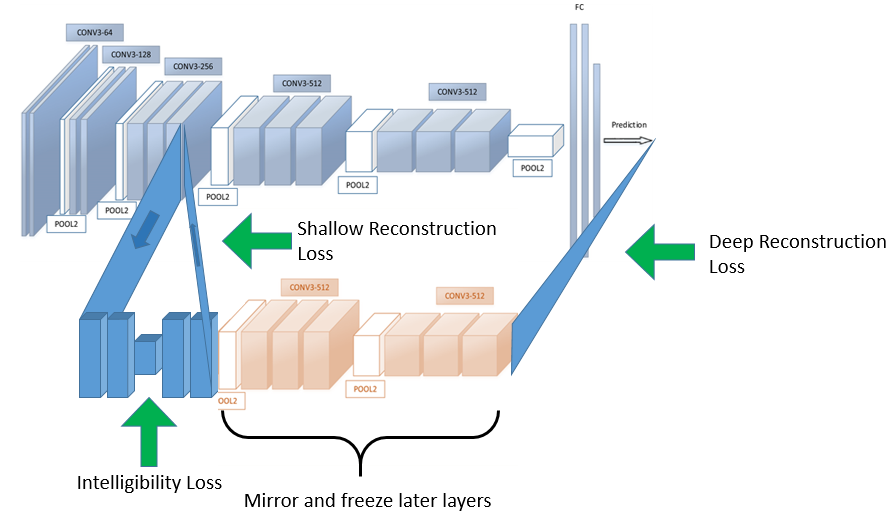}
  \caption{A schematic depiction of training autoencoders for activation compression. This depicts the architecture used for training an autoencoder for a single layer.}
  \label{fig:autoenc}
  \vspace{-10px}
\end{figure}

\begin{figure}
  \centerfloat
  \includegraphics[width=1.0\linewidth]{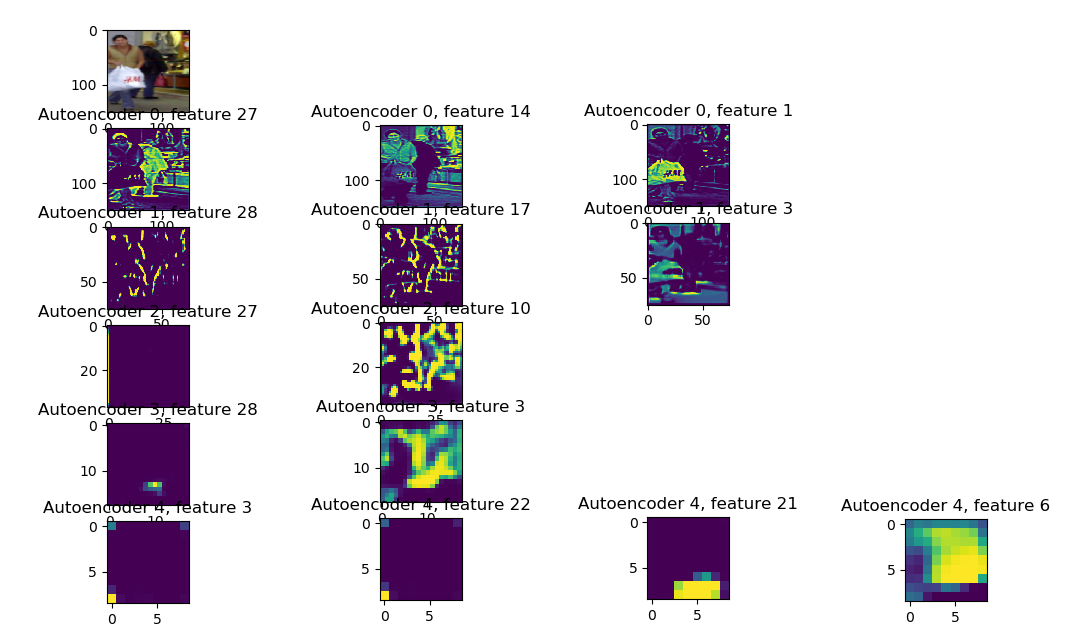}
  \caption{Sample display of coded features for an instance from VGG16 trained on the Inria pedestrian dataset. In the top left corner is the original input image. Each row corresponds to the extracted coded feature images from a different autoencoder, while each column corresponds to a different extracted feature image from that autoencoder.}
  \label{fig:inria_features}
  \vspace{-10px}
\end{figure}

\begin{figure}
  \centerfloat
  \includegraphics[width=1.0\linewidth]{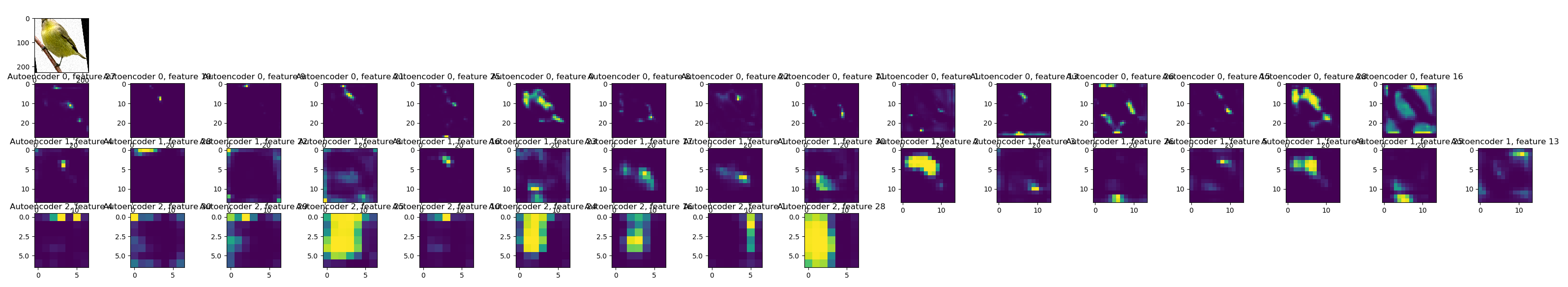}
  \caption{Sample display of coded features for an instance from VGG19 trained on birds200.}
  \label{fig:birds3}
  \vspace{-10px}
\end{figure}

The weighting hyperparameters were chosen by inspecting results and tuning by factors of 10 until output seemed reasonable - this only took 1 or 2 iterations of manual refinement per dataset. We trained autoencoders in this manner on activations at multiple layers throughout the network. The deep reconstruction loss is of particular benefit at the shallower layers, where one might not expect to be able to fit a linear classifier based on simple edge detectors. The training process proceeds by first training an autoencoder for the shallowest desired layer, then inserting the trained autoencoder into the network and training the next deepest autoencoder, iterating until all autoencoders have been trained. For our experiments we train 3-4 autoencoders spaced evenly throughout the convolutional layers of the target network. Each autoencoder consists of 3 convolutional layers in the coding and decoding networks. See Figure \ref{fig:autoenc} for a depiction of the architecture used for this training. See Figures \ref{fig:inria_features} and \ref{fig:birds3} for depictions of the resulting coded activations for sample input image instances.

Having trained these autoencoders, we now have a plausible approach for intervening on our network. If one were to intervene directly on network activations to probe its function it would be difficult to maintain the complex correlational statistics between the various components layer activations. Violating these statistics could result in activation values that would be impossible to recreate with any input sample, resulting in misleading output. In the autoencoder, on the other hand, we expect that correlations between features in the network activations would be captured by the encoding network. Intervening on the resulting code should ensure that decoded activations retain their statistics, as well as greatly reduce the size of possible interventions that one might consider. For our experiments, since we only autoencode with a few convolutional layers and do not fully encode all the way to a vector valued distribution, we restrict interventions to zeroing out individual concept feature images of the coded activations.

\section{Experiments}
Having trained autoencoders throughout the network, we now have a set of concepts we can intervene on (by changing the autoencoders code) and a known causal structure representing the relationship of these concepts. Given this, we construct a causal model describing the relationships between our concepts and the output prediction via known methods \cite{pearl2009causality}. In order to fit our causal model, we construct a large synthetic dataset containing some training set of input images and the values of their concept variables. We also randomly intervene on coded images in the autoencoders by zeroing out the entire feature image. This has a causal effect on all downstream layers that is captured in the pooled coded values of downstream variables. We intervene in this way on each coded feature image independently randomly with probability 0.1 and record the resulting values. We identify active coded feature images through a simple variance threshold, as many of them are always zero as a result of the sparsifying loss terms. These coded feature images are finally mean-pooled and binned into $k$ finite bins. We mean-pool concept images to make the construction of our bayes nets tractable - in future work we intend to improve on this approach. We found 2 bins were sufficient to maximize the probability of the data under our model according to these techniques. The feature values are then treated as variables in a causal Bayes net where each layer of autoencoded variables is dependent only on the variables of the previous layer. The shallowest autoencoded layer is treated as causally dependent on the class label as well as any other labels associated with the data instance.

\begin{figure}[t!]
  \centerfloat
  \includegraphics[width=0.8\linewidth]{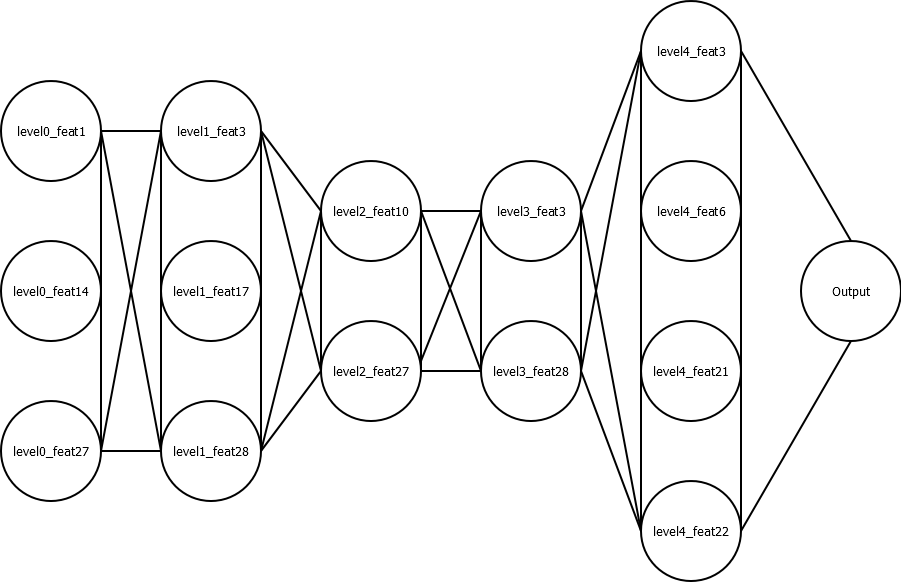}
  \caption{A graphical depiction of our learned causal Bayes net for JN6 applied to the Inria pedestrian dataset. The crossed boxes serve to indicate that all nodes of a given level have edges incident on each of the nodes of the subsequent level.}
  \label{fig:causal_model}
  \vspace{-10px}
\end{figure}

\begin{figure}
  \centerfloat
  \includegraphics[width=1.15\linewidth]{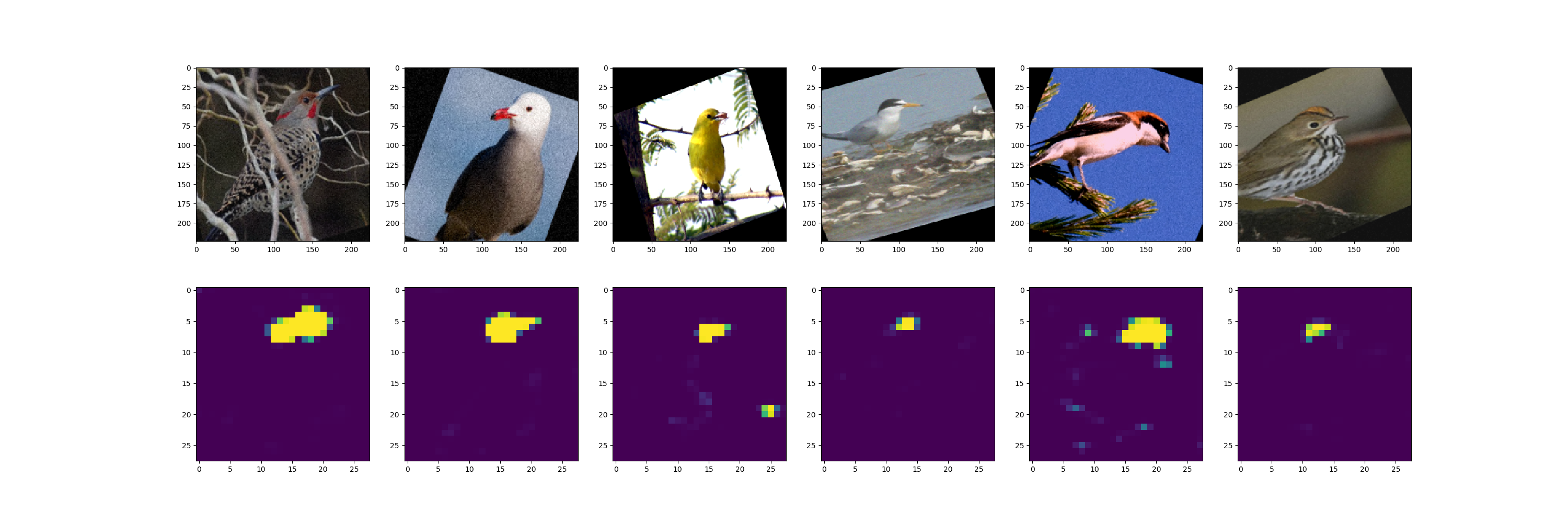}
  \caption{A resulting sample feature image displaying head identification on birds200 for the shallowest autoencoder (level0\_feat11) displayed alongside the nearest neighbors for that feature in a subset of an augmented dataset. The query image is the leftmost, as with the rest of these figures.}
  \label{fig:birds_nn1}
  \vspace{-10px}
\end{figure}

\begin{figure}
  \centerfloat
  \vspace{-10px}
  \includegraphics[width=1.15\linewidth]{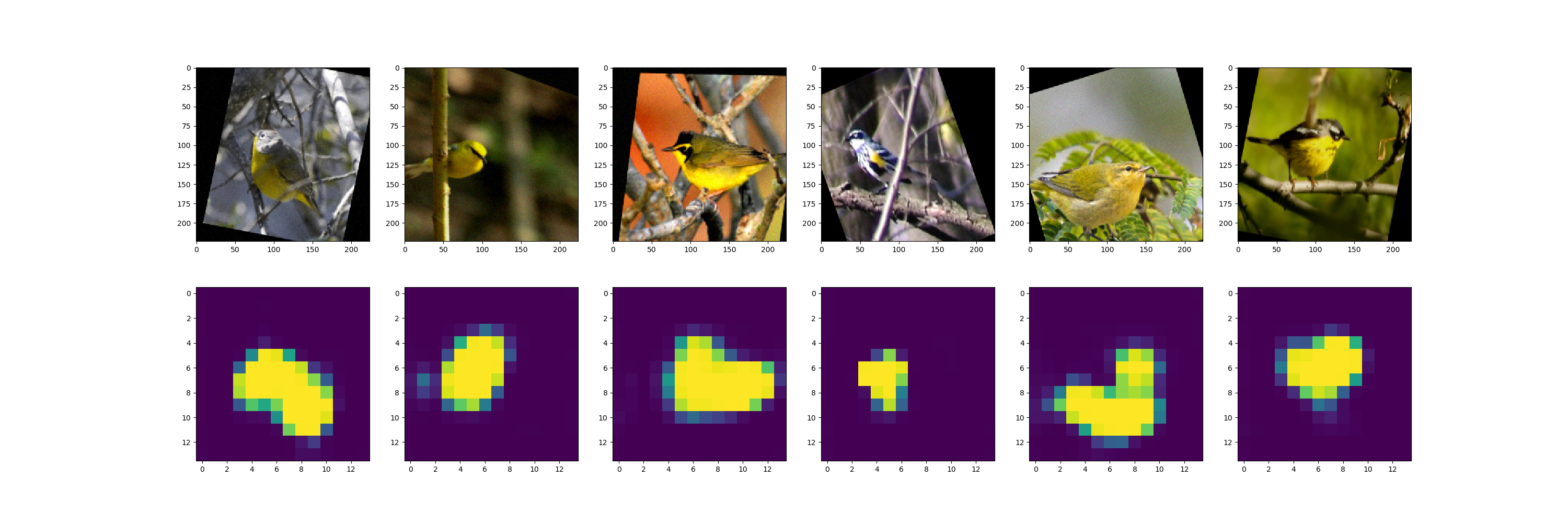}
  \caption{A resulting feature image displaying body color identification on birds200 for the 2nd autoencoder (level1\_feat2).}
  \label{fig:birds_nn2}
  \vspace{-20px}
\end{figure}

Then, having built a large synthetic dataset capturing which interventions were made and the causal effect of those interventions, we construct a bayes net as described above and fit the CPDs of each node to this dataset. See Figure \ref{fig:causal_model} for a graphical depiction of a sample learned causal model. With the resulting model we perform a query for individual input instances that ranks variables in the network according to their maximum causal effect on the output classification (see Eqn. 6). This is a single example of the types of causal queries that we can perform on this constructed model. These experiments were performed on 3 network architectures / datasets: (1) VGG 19 applied to Birds200 \cite{birds200}, (2) VGG 16 applied to the Inria pedestrian dataset \cite{inria}, and (3) A small 6 layer conv net we refer to as JN6 also applied to the Inria dataset.

\begin{figure}
  \centerfloat
  \vspace{-5px}
  \includegraphics[width=1.15\linewidth]{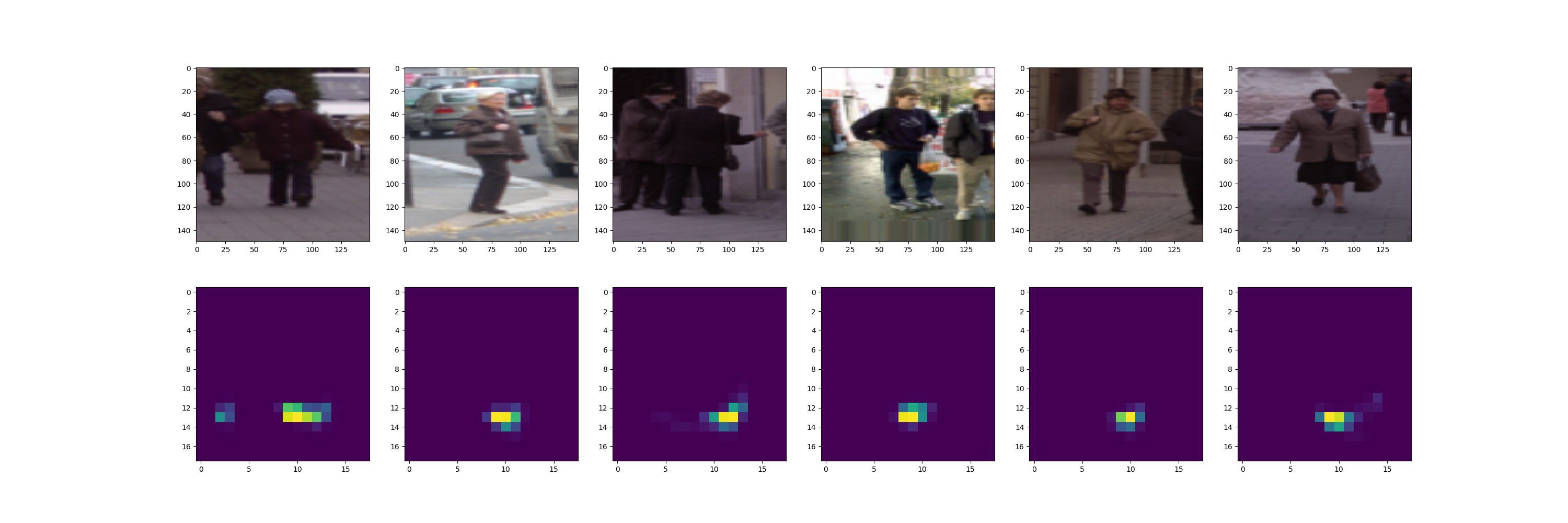}
  \caption{A feature image from the 4th autoencoder trained on vgg16 (level3\_feat28) applied to the Inria dataset depicting the identification of feet. This concept image has a very low average causal impact on the final classification (see Figure \ref{fig:expectedcausal}), indicating that visibility of feet may not have a major impact on classification.}
  \label{fig:inria_nn1}
  \vspace{-15px}
\end{figure}

\begin{figure}
  \centerfloat
  \includegraphics[width=1.15\linewidth]{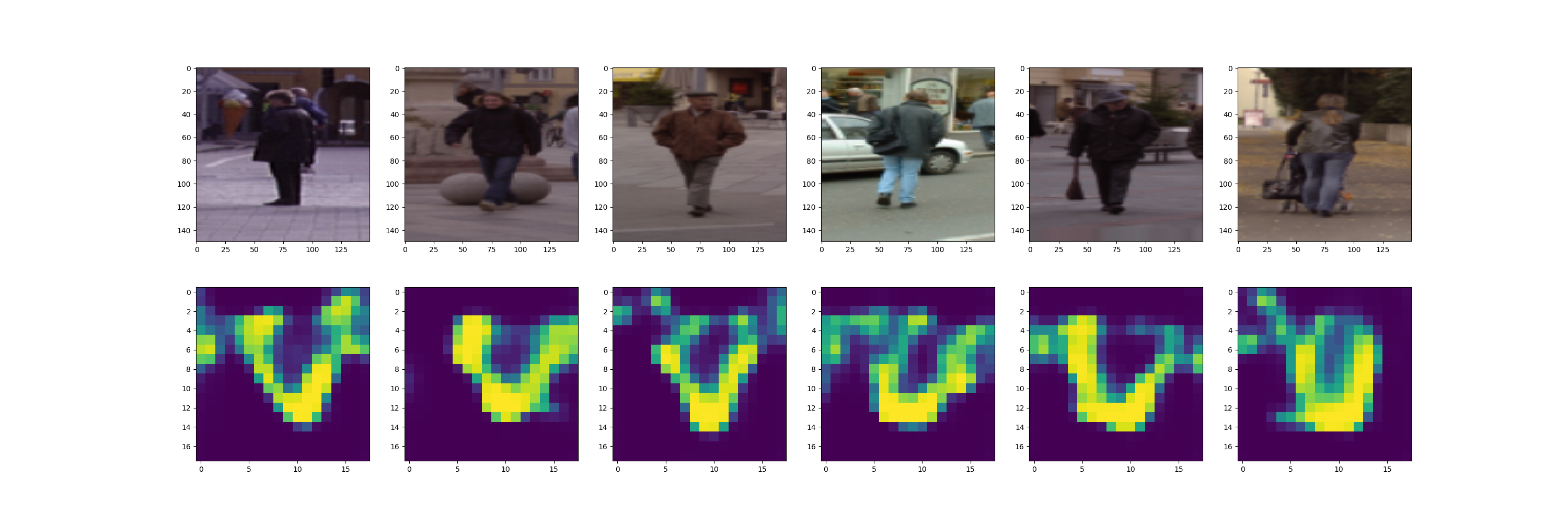}
  \caption{A feature image from the 4th autoencoder trained on vgg16 (level3\_feat3) applied to the Inria dataset depicting the identification of a person's outline. This feature has a large expected causal effect on the output of 0.097. On this input instance, this concept image (level3\_feat3) has an individual causal effect of 0.13, the largest causal effect of any concept images on this instance.}
  \label{fig:inria_nn2}
  \vspace{-15px}
\end{figure}

We then visualize the top $k$ variables according to their expected causal effect by displaying images along with the corresponding coded feature image. We additionally visualize the nearest neighbors in the dataset according to $l_1$ distance between these concept feature images, i.e. $|C_i^j-C_i^k|_1$ for the specified concept feature image $C_i$ and input instances $j$ and $k$. This helps the user to better interpret the feature image. See Figures \ref{fig:birds_nn1},\ref{fig:birds_nn2},\ref{fig:inria_nn1} and \ref{fig:inria_nn2} for instances of this nearest neighbor visualization. The final goal should enable a user to interrogate an input image instance of interest by automatically identifying concepts in the network that are highly relevant to classification (as measured by causal effect) and then visualize them in the context of other instances that contain that concept in a similar manner.

In Figure \ref{fig:expectedcausal} we list the expected causal effect over a dataset for VGG16 trained on the Inria pedestrian dataset. In addition to the depicted average causal effect table, we can query the causal model for individual classification instances. In this way we can identify the feature images with the maximum causal effect for the instance in question and analyze what they represent through nearest neighbor queries. This could be a highly useful tool for, for instance, debugging misclassifications in a DNN. As an example, in Figure \ref{fig:inria_nn2} we depict the concept feature image with the largest causal effect on that instance, which is level3\_feat3 with an individual causal effect of 0.13. We intend to develop an interactive tool to enable queries of this type and enable explanation of instances of interest. We omit additional specific results due to lack of space, though we intend to include more in a future release of this work.

\begin{figure}
	\begin{tabular}{l | r}
  Variable	& Expected Causal Effect \\ \hline
 level4\_feat6 & 0.174704302 \\
  level3\_feat3 & 0.09731648 \\
  level2\_feat10  & 0.056770524 \\
  level1\_feat3 & 0.028265387 \\
  level1\_feat17  & 0.023817493 \\
  level0\_feat27  & 0.016577831 \\
  level2\_feat27  & 0.01370528 \\
  level0\_feat1 & 0.0123624 \\
  level4\_feat3 & 0.007728 \\
  level4\_feat22  & 0.007587164 \\
  level0\_feat14  & 0.006091733 \\
  level4\_feat21  & 0.002876711 \\
  level1\_feat28  & 0.001066667 \\
  level3\_feat28  & 7.24E-04
	\end{tabular}
  \caption{The resulting expected causal effect query across the entire dataset applied to VGG16 on Inria . 'Level' denotes which autoencoder (the shallowest being level0) and 'feat' indicates which coded feature image channel it refers to (of those that are active after pruning).}
  \label{fig:expectedcausal}
  \vspace{-15pt}
\end{figure}
\label{results}

\section{Conclusion}
\label{conclusion}

To summarize, we describe an approach to explaining the predictions of deep neural networks using causal semantics to relate the output prediction of a network to concepts represented within. We use a series of autoencoders with loss functions encouraging interpretable properties to construct concepts representing the information content in activations throughout a target network. These autoencoders are trained with a novel "deep" loss that allows increased flexibility in representation. We pool these features and intervene on our autoencoded network to construct variables that we use to build a causal bayesian network which fits the causal relationship defined by the network structure. We finally use this network to identify features of significant causal relevance to individual classifications which are then visualized via our described approach.

This an early investigation of ideas in this domain. There are a number of interesting possible directions for future work. One clear area of potential improvement is the use of more sophisticated methods to construct variable observations for our bayes net. In the future we intend to explore the construction of variational autoencoders where all image structure is encoded away to allow for the compression of irrelevant image structure. This would greatly increase the size of bayes nets (in terms of incident edges on nodes), which suggests that it may be prudent to consider structure learning for reducing the size of the bayes net skeleton. Additionally, we'd like to consider the causal relationship between rich input labels, network concept features and ultimate classification. This could enable direct identification of the parts of the network that identify relevant input concepts (eg. what part of the network detects heads?) and how those components contribute to ultimate classification, as well as direct identification of confounding concepts that could result in incorrect classification (eg. classifications are often incorrect when it is dark out). Finally, we are interested in extending these approaches to non-image domains.

\section*{Acknowledgments}
This material is based upon work supported by the United States Air Force under Contract No. FA8750-17-C-0018 for the DARPA XAI program. Distribution A. Approved for public release. Distribution is unlimited.

\bibliographystyle{named}
\bibliography{AE_Explain}

\end{document}